\documentclass[runningheads]{llncs}
\usepackage[hidelinks]{hyperref}
\usepackage{graphicx}
\usepackage{algorithm}
\usepackage{algpseudocode}
\begin{document}
\title{ Unsupervised Domain Adaptation in Semantic Segmentation Based on
Pixel Alignment and Self-Training (PAST)}
\titlerunning{PAST}
%
\author{Hexin Dong \inst{1*}  \and
Fei Yu \inst{1*} \and
Jie Zhao\inst{3} \and
Bin Dong \inst{4,1} \and
Li Zhang \inst{1,2}\thanks{Correspondence to: zhangli\_pku@pku.edu.cn}
}

\authorrunning{Dong et al.}
%
\institute{Center for Data Science, Peking University, Beijing, China\and
Center for Data Science in Health and Medicine, Peking University, Beijing, China\and
National Engineering Laboratory for Big Data Analysis and Applications, Peking University, Beijing, China\and
Beijing International Center for Mathematical Research (BICMR), Peking University, Beijing, China ~~~~*These authors contributed equally
\email{\{donghexin,yufei1900,jiezhao,zhangli\_pku\}@pku.edu.cn}
dongbin@math.pku.edu.cn
}

\maketitle              
\begin{abstract}

This paper proposes an unsupervised cross-modality domain adaptation approach based on pixel alignment and self-training. Pixel alignment transfers ceT1 scans to hrT2 modality, helping to reduce domain shift in the training segmentation model. Self-training adapts the decision boundary of the segmentation network to fit the distribution of hrT2 scans. Experiment results show that PAST has outperformed the non-UDA baseline significantly, and it received rank-2 on CrossMoDA validation phase Leaderboard with a mean Dice score of 0.8395.
\end{abstract}
\section{Introduction}
CrossModa challenge\cite{Shapey2021.08.04.21261588,Shapey:SDATA:2021,Shapey2021_data} aims to segment two critical brain structures involved in the treatment planning of vestibular schwannoma (VS): the tumor and the cochlea.  While contrast-enhanced T1 (ceT1) Magnetic Resonance Imaging (MRI) scans are commonly used for VS segmentation, recent work has demonstrated that high-resolution T2 (hrT2) imaging could be a reliable, safe, and lower-cost alternative to ceT1. Therefore, the participants are asked to provide a segmentation model of VS and cochlea on hrT2 scans based on unsupervised domain adaptation (UDA) using only the information of labeled ceT1 scans and unlabeled hrT2 scans.

To solve this problem, we propose an effective and intuitive UDA method combining pixel-level alignment and self-training (PAST). Firstly, we transfer labeled images from the ceT1 domain to the hrT2 domain so that images can be aligned into the same distribution. Secondly, the model is further trained on pseudo labels generated from transferred ceT1 scans and hrT2 scans, which find a better decision boundary on the hrT2 domain. The experimental results show that our method greatly reduces the domain shift and achieves 2nd place with a dice score of 0.8395 on the validation set.

\section{Methods and Experimental Methods}
\subsection{Method Overview}
We introduce our method in this section. Our method has two major parts: pixel-level alignment and the self-training stage. 

\begin{figure}
\centering
\includegraphics[width=0.9\textwidth]{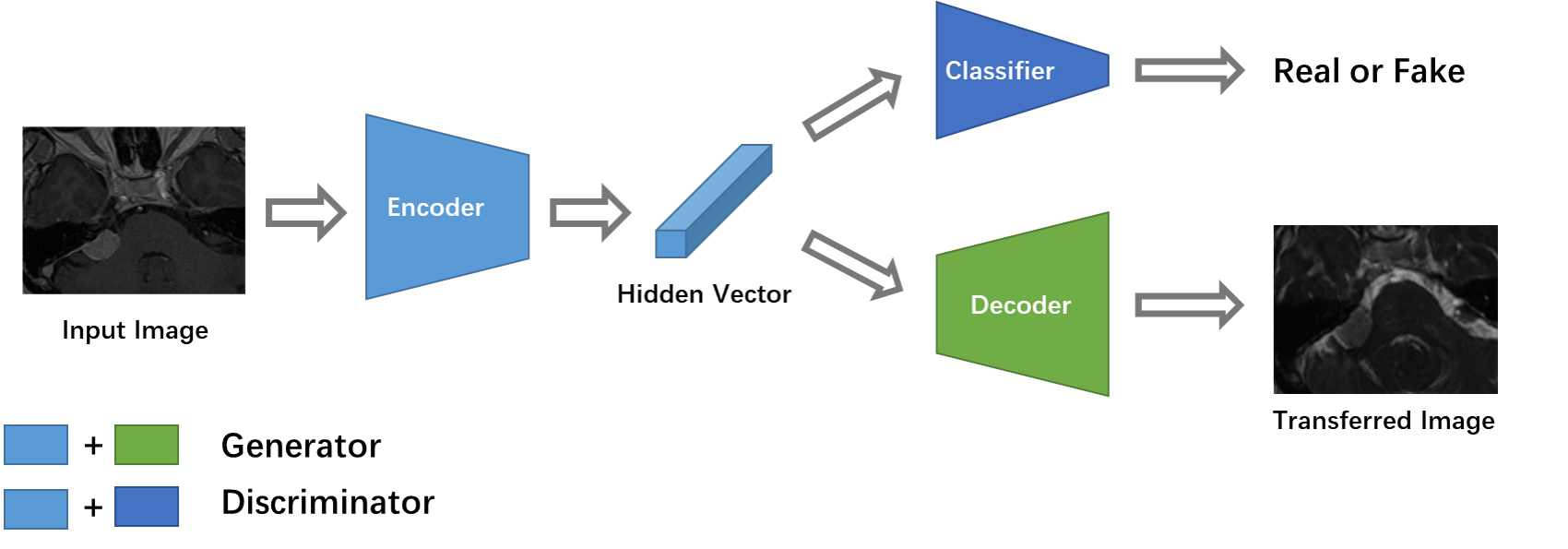}
\caption{flowchart of NiceGAN\cite{chen2020reusing}. It extracts feature from the input image with the shared Encoder. The Classifer from the Discriminator distingguishes  real or fake feature vectors. The Decoder from the Generator generates transferred images.
} 
\label{fig-nicegan}
\end{figure}

First, we follow \cite{hoffman2018cycada} to learn a mapping from the source domain to the target domain, i.e., we transfer ceT1 scans to hrT2 scans. After doing so, we can use synthesized hrT2 scans to train a segmentation model using supervised learning. As shown in \autoref{fig-nicegan}, the model achieves the domain adaptation using NiceGAN \cite{chen2020reusing} (i.e., an extension method of CycleGAN), which reuses discriminators for encoding to improve the efficiency and effectiveness of training.

Second, we apply self-training to further improve the decision boundary of the segmentation model. Similar to \cite{li2019bidirectional}, we introduce a super parameter $q$ of the pixel portion. We iteratively generate the pseudo label $\hat{y_c}$ using the top $q$ of pixels in segmentation output $y_c$ with a higher probability to retrain the model. Overall training process of the proposed method is summarized in Algorithm \ref{alg-1}.

All models are implemented using the PyTorch 1.9. Pixel-level alignment model
runs on a single V100 GPU with 16 GB memory and self-training model runs on a single TIAN V GPU with 12 GB memory. All training data used are collected from CrossModa training set \cite{Shapey2021_data,Shapey2021.08.04.21261588} and we verify our model on crossModa validation set.
\begin{algorithm}[!h]
\caption{training process of the proposed method}
\label{alg-1}
\begin{algorithmic}[1]
\State Initialize ceT1 scans images and label $(X_s,y_s)$, hrT2 scans images $X_t$, Segmentation network $S$, Image translation network $T$
\State Train network $T$ with $X_s$ and $X_t$ 
\State Transfer ceT1 scans $X_s$ to $\hat{X_s}$ using $T$
\State Train network $S$ with $(\hat{X_s},y_s)$
\State Initialize concat scans images $X_c=\{\hat{X_s},X_t\}$, self-training segmentation network $S_0=S$
\For{$k \gets 1$ to $K$}
\State input $X_c$ into $S_{k-1}$ and generate pseudo label $\hat{y^k_c}$ with a fixed portion $q_k$
\State Initialize $S_{k} \gets S_{k-1}$
\State Train $S_{k}$ with $(X_c,\hat{y^k_c})$
\EndFor
\State \Return $S_k$
\end{algorithmic}
\end{algorithm}

\subsection{Experiments}
\begin{figure}[htb]
\centering
\includegraphics[width=0.7\textwidth]{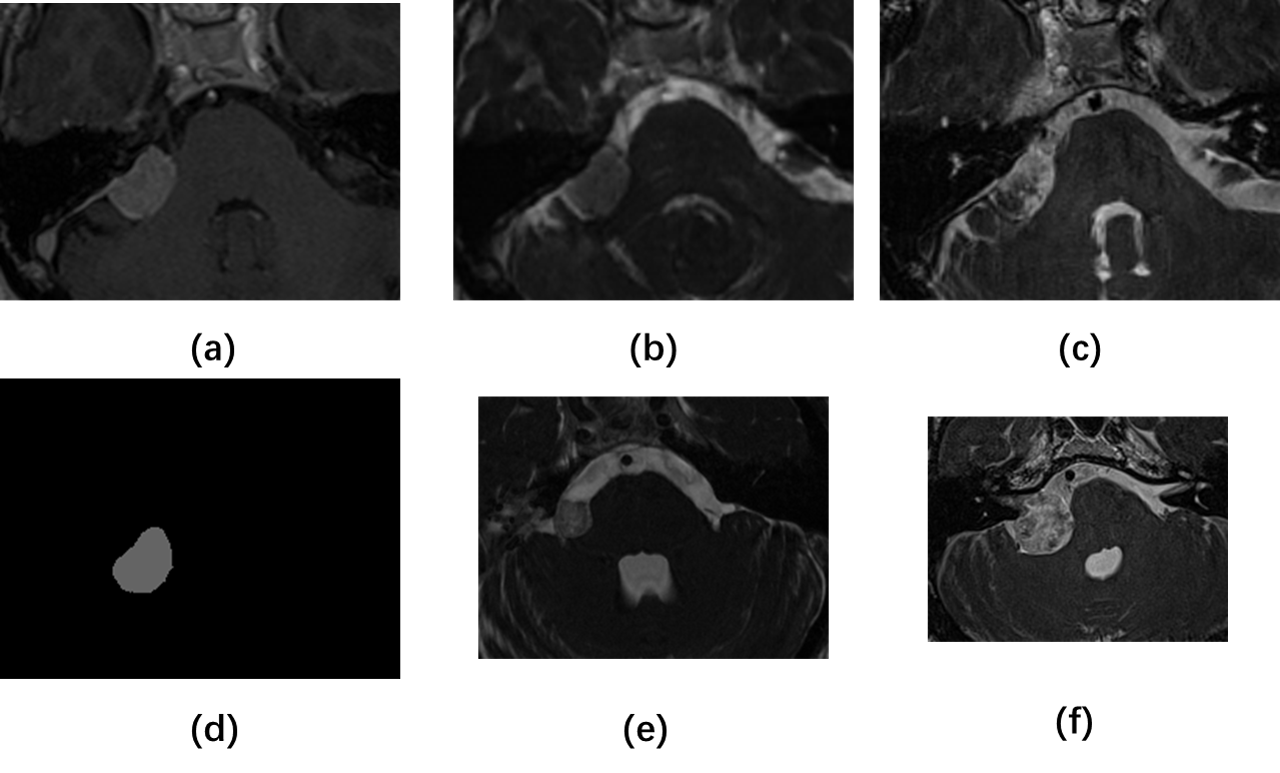}
\caption{Croped images samples: a).ceT1 sample b).transfered 448 ceT1 sample c).transfered 384 ceT1 sample d).label e).448 hrT2 sample f).384 hrT2 sample} 

\label{fig1:crop}
\end{figure}

For the preprocessing step, we observe that the segmentation targets are located in the center of the image, so we take the center area of the image as the region of interest (ROI) (Figure~\ref{fig1:crop}). For the image translation, the 2D images with a range of $[0:W, 0:H]$ are cropped into the 2D ROI with a range of $[\frac{W}{4}:\frac{3W}{4}, \frac{H}{4}:\frac{3H}{4}]$. For the 3D segmentation, the volumetric data ($[0:W, 0:H, 0:D]$) will be cropped into the 3D ROI with a range of $[\frac{W}{4}:\frac{3W}{4}, \frac{3H}{8}:\frac{3H}{4}, 0:D]$. After then, the intensity values in ROI are normalized by rescaling to $[0-255]$ 

In the pixel alignment stage, we adopt NICE-GAN \cite{chen2020reusing} on 2D transverse slides of the ROIs, transferring ceT1 to hrT2. We then concatenate the synthesized 2D hrT2 slides to a 3D volumetric image. For 3D segmentation, we follow the nnUNet framework \cite{isensee2019automated}. Several research settings are implemented. First, we train the models with paired synthesized hrT2 scans and labels. Since most data have a dimension of 448 pixels, we thus call this model \textbf{nnUnet448}. However, we notice that there’re two types of protocols in hrT2 with significantly different appearances (one with a dimension of 448 pixels, called 448 scans, and another with a dimension of 384 pixels, called 384 scans). We thus create a \textbf{nnUnet384} model for those 384 scans. We evaluate the results of the two models on all data and the results of their respective applicable data. We call the model in the latter scenario as \textbf{nnUnetCon}.

The self-training stage generally follows the Algorithm \ref{alg-1}. The nnUnets\\ (\textbf{nnUnet448}, \textbf{nnUnet384} and \textbf{nnUnetCon}) are used as pretrain models for self-training, i.e. $S_0$ in Algorithm \ref{alg-1}. The synthesized images and pseudo labels are derived from the corresponding image-to-image translation models and nnUnets respectively. In practice, we set the initial $q$ to $0.6$, the maximum iteration $K$ of self-training to 2, and the initial learning rate is set to $0.08$. We name this model as \textbf{nnUetST2}.

Apart from this, we train a ResUnet (i.e., an extended version for nnUnet with ResNet encoder) following the above stages and name this model as \textbf{ResUnetST2}. A combined version using \textbf{nnUetST2} to segment cochlea and \textbf{ResUnetST2} to segment VS has also been evaluated and achieves a better result. We thus name it the proposed \textbf{PAST}.

\section{Results}
\begin{figure}[htb]

\centering
\includegraphics[width=0.8\textwidth]{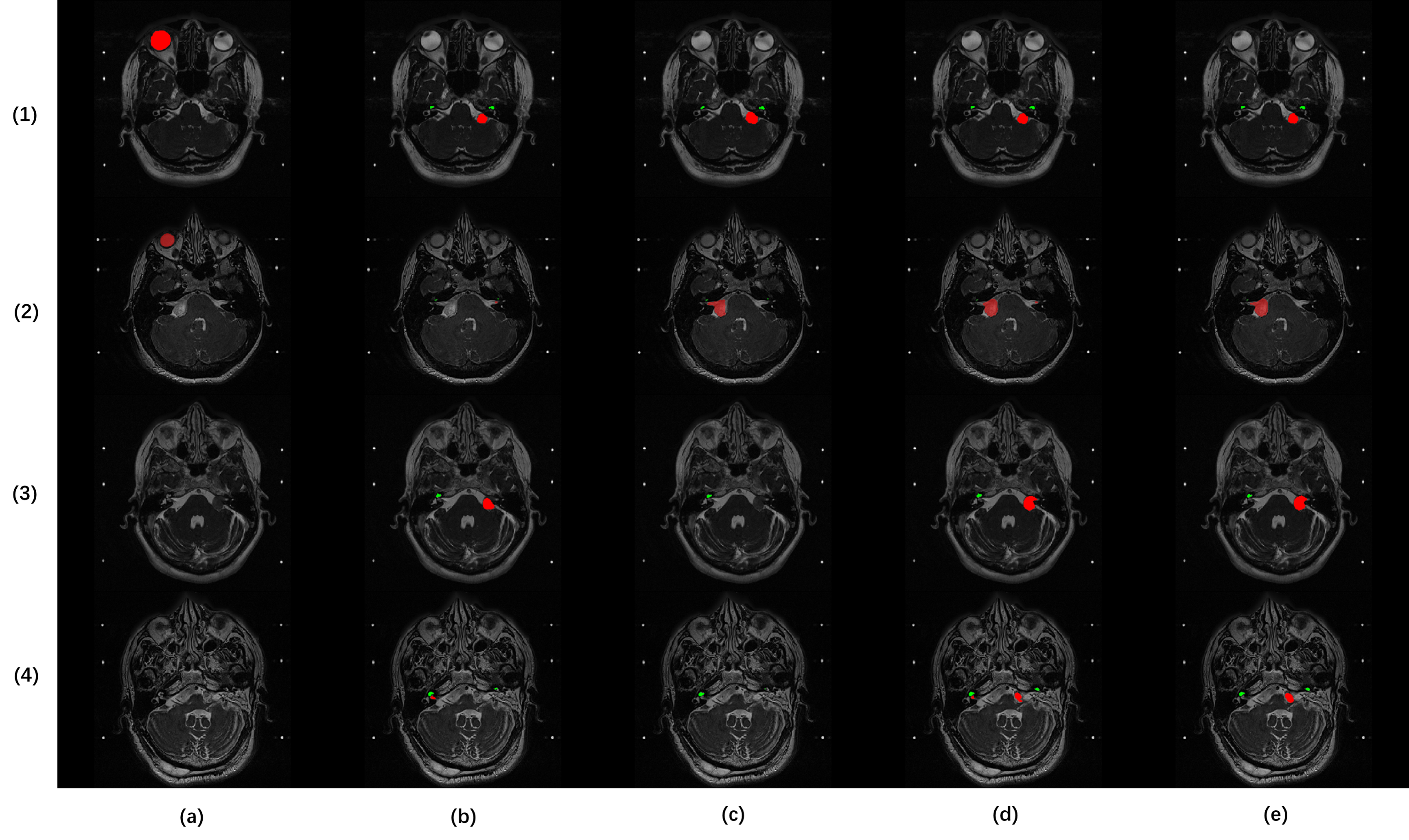}
\caption{Visual results of segmentation output : a).nnUnet without DA b).nnUnet448 c).nnUnet384 d).nnUnetST2 e).ResUnetST2 ;
Sample id : 1).211(448) 2).213(384) 3).214(448) 4).240(384)} 
\label{fig1:result}
\end{figure}

Table \ref{tab:result} shows the results for the aforementioned models. Compared to nnUnet without DA, i.e., training with ceT1 scans directly, both \textbf{nnUnet448} and \\ \textbf{nnUnet384} have a noticeable improvement, which shows the effectiveness of the pixel alignment. However, the two models did not achieve satisfactory accuracy because hrT2 modality itself has two different protocols. \textbf{nnUnetCon} model solves this problem and further improves the performance with model ensembling. Furthermore, the experiments show that self-training achieves better performance on overall Dice, but different network backbones behave differently on either VS or cochlea segmentation (Figure~\ref{fig1:result}). Thus, as our final proposed method, PAST combines all the merits from the aforementioned network architectures and training techniques, raising the overall Dice to 0.8395.
\begin{table}[htb]
\centering
\caption{Segmentation results for selected model.}

\begin{tabular}{|c|c|c|c|}

\hline
Model Name        & VS Dice & Cochlea Dice & Mean Dice \\ \hline
nnUnet without DA &  $ 0.0549 \pm 0.1859 $       &   $0.1905 \pm 0.1643 $           &    $0.1227 \pm 0.1192 $       \\ \hline
nnUnet448 & $0.7509 \pm 0.2683 $&     $0.7818 \pm 0.0425 $ &   $0.7664 \pm 0.1417$        \\ \hline
nnUnet384         &   $ 0.5905 \pm 0.3754 $      &     $0.7870 \pm 0.0413 $ & $0.6887 \pm 0.1929 $  \\ \hline
nnUnetCon &   $0.8281\pm0.1679$      &   $0.7949 \pm 0.0332 $   &        $0.8115\pm0.0848$   \\ \hline
nnUnetST2& $ 0.8553 \pm 0.0871$ &    $0.8089 \pm 0.0334 $&  $ 0.8321 \pm 0.0435 $ \\ \hline
ResUnetST2        &   $0.8700 \pm 0.0657$      &     $0.7820 \pm 0.0295$         &  $0.8260 \pm 0.0349$ \\ \hline
PAST        &  $0.8700 \pm 0.0657$   &  $0.8089 \pm 0.0335$   & $0.8395 \pm 0.0328$
\\ \hline
\end{tabular}
\label{tab:result}
\end{table}

%
%
%

\end{document}